\def\year{2022}\relax
\documentclass[letterpaper]{article} 
\usepackage{aaai22}  
\usepackage{times}  
\usepackage{helvet}  
\usepackage{courier}  
\usepackage[hyphens]{url}  
\usepackage{graphicx} 
\usepackage{natbib}  
\usepackage{caption} 
\DeclareCaptionStyle{ruled}{labelfont=normalfont,labelsep=colon,strut=off} 
\usepackage{algorithm}
\usepackage{algorithmic}
\usepackage{amsmath}
\usepackage{tabularray}
\usepackage{newfloat}
\usepackage{listings}
\floatstyle{ruled}
\newfloat{listing}{tb}{lst}{}
\floatname{listing}{Listing}

\title{Implementing Long Text Style Transfer with LLMs through Dual-Layered Sentence and Paragraph Structure Extraction and Mapping}
\author {
    Yusen Wu,\textsuperscript{\rm 1}
    Xiaotie Deng, \textsuperscript{\rm 1}
}
\affiliations {
    \textsuperscript{\rm 1} School of Computer Science, Peking University\\
}

\usepackage{bibentry}

\begin{document}

\maketitle

\begin{abstract}
This paper addresses the challenge in long-text style transfer using zero-shot learning of large language models (LLMs), proposing a hierarchical framework that combines sentence-level stylistic adaptation with paragraph-level structural coherence. We argue that in the process of effective paragraph-style transfer, to preserve the consistency of original syntactic and semantic information, it is essential to perform style transfer not only at the sentence level but also to incorporate paragraph-level semantic considerations, while ensuring structural coherence across inter-sentential relationships. Our proposed framework, ZeroStylus, operates through two systematic phases: hierarchical template acquisition from reference texts and template-guided generation with multi-granular matching. The framework dynamically constructs sentence and paragraph template repositories, enabling context-aware transformations while preserving inter-sentence logical relationships. Experimental evaluations demonstrate significant improvements over baseline methods, with structured rewriting achieving 6.90 average score compared to 6.70 for direct prompting approaches in tri-axial metrics assessing style consistency, content preservation, and expression quality. Ablation studies validate the necessity of both template hierarchies during style transfer, showing higher content preservation win rate against sentence-only approaches through paragraph-level structural encoding, as well as direct prompting method through sentence-level pattern extraction and matching. The results establish new capabilities for coherent long-text style transfer without requiring parallel corpora or LLM fine-tuning.
\end{abstract}

\section{Introduction}
Text Style Transfer (TST) focuses on modifying specific stylistic attributes of text while preserving its underlying content \cite{def}. The task aims to adapt texts to meet diverse stylistic criteria—such as sentiment, formality, or politeness—without altering their explicit meaning. This capability plays a vital role in enhancing communication effectiveness and refining writing quality, particularly in scenarios requiring stylistic adaptation (e.g., rendering text more polite or contextually appropriate for formal settings). In academic writing, where stylistic variations across authors can hinder clarity, TST offers significant utility. By adjusting academic tone to improve positivity or removing inappropriate language, it facilitates author interactions and mitigates misinterpretations. Formally, TST rephrases text to incorporate new or alternative stylistic elements while retaining semantic and structural fidelity \cite{def}. Applications span diverse use cases, including \cite{case1}, \cite{case2}, and \cite{case3}.

The research on text style transfer has seen significant evolution with the advent of natural language processing (NLP) techniques. Early studies primarily focused on sentence-level stylistic modeling. For instance, \cite{nlp2stages} proposed a two-stage generation framework that disentangles content planning from stylistic control for paragraph-level text generation, though input remained limited to topic statements of hundreds of words. With the emergence of unsupervised learning, researchers developed probabilistic models for single-sentence style transfer \cite{nlp_sentence} or optimized word-level stylistic features through discrete editing strategies \cite{prompt-edit}. While these methods achieved progress at the sentence level, they face challenges in ensuring coherence in long-text generation.

The emergence of large language models (LLMs) has introduced a paradigm shift in style transfer. Their generative capabilities enable both zero-shot and fine-tuning transfer methods. Current research primarily follows two directions: First, stylistic adaptation in dialogue scenarios, exemplified by \cite{lmstyle-bench}'s LMStyle Benchmark framework, which introduces appropriateness metrics to evaluate style consistency in chatbot responses. Second, model fine-tuning strategies, such as \cite{attn-mask}'s approach using attention masking and multi-path interaction to enhance sentence-level transfer. Notably, recent work has begun investigating document-level style conversion, as demonstrated by \cite{catllm}'s CAT-LLM system for Chinese article style definition and transfer. However, such methods still rely on domain-specific parallel data and require significant computational resources. In unsupervised scenarios, while \cite{prefix-tune} improved single-sentence transfer through prefix tuning and \cite{sim-real-person} employed in-context learning for author style imitation, their processing units remain limited to short texts.
Current zero-shot long-text style transfer faces two core challenges: First, existing methods are typically limited to single-sentence or single-turn dialogue conversion. When extended to document-level tasks, they exhibit style persistence degradation. As observed in dialogue style transfer \cite{conversation, cotex}, models demonstrate style drift during multi-turn processing. Second, current evaluation systems inadequately capture macro-stylistic features. Although \cite{textsettr} proposed paragraph-level style adjustment through style vector extraction, their metrics remain confined to lexical similarity measurements. They fail to assess inter-sentence logical coherence and other deep stylistic elements. This limitation arises because traditional approaches treat style as local feature aggregations while neglecting the role of text structure as a style carrier, such as authors' paragraph development patterns and argumentation logic sequences \cite{author-style, sim-real-person}. 

Consequently, there is a critical need for systematic style parsing frameworks that jointly model micro-linguistic features and macro-structural patterns to establish long-text-adaptive transfer paradigms.
Most effective style transfer methods depend on model training or fine-tuning requiring massive stylistic corpora (e.g., an author's complete works dataset)\cite{finetune1,finetune2}, which are often unavailable for specific tasks and entail high computational costs. Meanwhile, current LLM-based zero-shot style transfer approaches, despite inspiring progress, predominantly focus on sentence-level conversion with limited exploration of long-text scenarios. The long-text style conversion problem manifests particularly in LLM applications. When processing lengthy inputs, models frequently exhibit premature termination of style adaptation. Beyond certain length thresholds, they selectively modify partial paragraphs despite explicit conversion instructions. A potential solution involves text segmentation for sequential processing using mature sentence-level techniques, but this risks losing inter-sentence structural and stylistic coherence. As stylistic analysis reveals, writing style encompasses not only sentence-level expressions but also paragraph relationships and logical sequencing—both critical stylistic components \cite{catllm}.

To address these challenges, we propose a zero-shot hierarchical framework for long-text style transfer using LLMs. Our approach systematically combines sentence-level stylistic adaptation with paragraph-level structural coherence through a two-stage process. During style abstraction, the framework extracts expression patterns from reference style paragraphs, constructs reusable templates at both sentence and paragraph levels, and dynamically matches these templates to guide text rewriting. The methodology specifies three key phases: First, sentence templates are extracted by parsing reference texts to identify recurring logical expressions, which are de-duplicated and organized into a template repository. These sentence templates are then mapped to paragraph-level patterns through clustering algorithms, forming hierarchical style representations. During rewriting, each sentence in the input text is processed sequentially using LLMs. Its logical structure is matched against the sentence template repository, and the framework identifies optimal paragraph templates that align with aggregated sentence patterns while preserving inter-sentence coherence.

A critical innovation lies in the decoupling of sentence and paragraph template mappings. This enables selective style adaptation using subsets of reference materials (e.g., temporal-specific paragraph templates), allowing dynamic style updates without reprocessing entire corpora. To mitigate LLM degeneration in long-text processing, we implement length-constrained iterative rewriting. Text segments are processed within bounded context windows, ensuring consistent style application while preventing premature termination of stylistic adjustments. The framework inherently addresses two fundamental requirements of long-text style transfer: (1) Preservation of paragraph-level structural patterns through template-guided rewriting sequences, and (2) Maintenance of micro-stylistic consistency via sentence-template alignment. Through experimental evaluations, we demonstrate superior style retention performance compared with baseline methods. Ablation studies confirm the necessity of both hierarchical template matching and length-constrained generation components.

\section{Related Work}
\subsection{Traditional Style Transfer}
Research on text style transfer has evolved from localized to holistic approaches and from supervised to unsupervised paradigms. Early efforts focused on sentence-level style conversion through content-style disentanglement\cite{tradition-tst1,tradition-tst2,tradition-tst3}. While these methods achieved strong performance on automatic metrics, they might be limited to single-sentence processing and missed to ensure coherence in long-text generation. To address the scarcity of parallel corpora, subsequent studies introduced contrastive learning strategies, leveraging back-translation and pseudo-parallel corpus construction to separate content and style representations \cite{textsettr}. However, these approaches displayed shortage in global awareness of text structure, usually leading to style fragmentation in paragraph-level transfers. The integration of adversarial learning with variational autoencoders attempted style-content disentanglement in latent spaces \cite{author-style}, yet struggled to capture explicit linguistic features, especially when handling Chinese-specific phenomena like classical vernacular style transfer \cite{catllm}. Here, multi-level preservation of lexical, syntactic, and cultural connotations posed significant challenges.

\subsection{Style Transfer with LLMs}
The emergence of large language models (LLMs) has transformed style transfer paradigms. Zero-shot prompting methods enable flexible style adaptation through instruction tuning and in-context learning \cite{prompt-edit}. Applications include enhancing response history knowledge in dialogue systems via retrieval-augmented mechanisms \cite{history} and guiding emotion-style transfer classifiers \cite{emotion}. Notably, while these approaches maintain style consistency across multi-turn dialogues, their evaluation systems predominantly rely on lexical similarity metrics (e.g., BLEU, self-BLEU, and perplexity) \cite{bleu,attn-mask,prefix-tune,self-bleu}, not fully covering quantitative analysis of macro-stylistic elements such as argumentation logic and paragraph development patterns. Recent explorations into document-level frameworks remain constrained by domain-specific parallel data requirements and struggle with long-range consistency in unsupervised settings.

Basically current research faces two fundamental challenges: long-text coherence preservation and evaluation system adaptation. Traditional methodologies treat style as discrete local feature collections, neglecting text structure's role as a style carrier. For instance, in dialogue style transfer, models exhibit style drift beyond multiple conversational turns \cite{conversation,cotex}, attributable to inadequate modeling of inter-sentence logical relationships and topic continuity. Partial solutions include hierarchical style parsing frameworks such as synergistic content planning and style control decoders \cite{nlp2stages} or attention masking mechanisms for enhanced multi-path interaction \cite{attn-mask}. However, these methods still suffer from selective paragraph modification in document-level transfers. Evaluation-wise, existing methods remain insufficient in capturing deep stylistic features like author-specific argumentation logic and rhetorical preferences, necessitating unified evaluation frameworks that integrate micro-linguistic features with macro-structural patterns.

\subsection{Zero-Shot Inference and Chain-of-Thought in LLMs}
Research on LLMs has increasingly emphasized inference-time optimization techniques, including few-shot and zero-shot learning, driven by the prohibitive computational demands and uneven resource distribution associated with pretraining and fine-tuning. These challenges hinder the fulfillment of diverse task requirements such as stylistic adaptation, personalized customization, and meta-domain applications \cite{coser,personalize,domain}. Consequently, scholars have explored methods to enhance model performance without architectural modifications or additional training, primarily through strategic prompt engineering. A pivotal advancement in this paradigm is Chain-of-Thought (CoT) \cite{cot}, which significantly improves problem diagnosis, iterative refinement, and reasoning extension capabilities. By decomposing complex tasks into multi-step reasoning processes—either through meticulously designed prompts or automated generated CoT enables LLMs to address errors incrementally and refine intermediate outputs. This approach effectively trades computational resources at inference time for enhanced final-output accuracy \cite{plan1, plan2} without parameter updates.

Current agent systems extensively leverage CoT-driven prompting strategies to achieve human-aligned task execution\cite{cot-agent1,cot-agent2}. These methodologies underpin state-of-the-art implementations in some settings where agents perform iterative environment analysis, stepwise plan formulation, and self-corrective action sequences. Such frameworks demonstrate particular efficacy in domains requiring contextual adaptation and meta-reasoning, aligning with the original goals of inference-time optimization for personalized and resource-efficient AI systems. In style transfer settings, prefix tuning \cite{prefix-tune} and self-explanatory distillation \cite{cotex} offer novel pathways to reduce data dependency. While achieving remarkable single-sentence transfer through chain-of-thought prompting, model capability distillation \cite{cotex} or few-shot learning \cite{conversation} still face persistent style degradation in long-text scenarios.

\section{Preliminaries}
\subsection{Adapting Language Models for Non-Parallel Author-Stylized Rewriting}
Stylized text generation remains a challenging task in NLP. \cite{author-style} proposes StyleLM, a method for rewriting input texts into author-specific stylistic variations without parallel data. StyleLM first pre-trains a Transformer-based language model on a large corpus and then fine-tunes it on a target author’s corpus via a cascaded encoder-decoder framework. A denoising autoencoder (DAE) loss function is incorporated to enable the model to capture stylistic features while preserving semantic content.

Experimental results demonstrate StyleLM’s superiority in style alignment compared to baselines, as validated by quantitative metrics (e.g., BLEU, ROUGE) and qualitative assessments. To evaluate performance, \cite{author-style} introduces a linguistically motivated framework that quantifies style alignment across three dimensions—lexical, syntactic, and surface—and measures content preservation using standard metrics. Style consistency is assessed via distance metrics such as mean squared error (MSE) and Jensen-Shannon divergence (JSD). This framework eliminates reliance on external classifiers, offering interpretable evaluations.

Despite these advances, StyleLM struggles with long sentences and complex style transfers. Empirical analysis shows the model excels on short texts and simple stylistic features but falters on lengthy passages or intricate patterns. These findings suggest architectural refinements and training optimizations are needed to improve handling of complex linguistic structures.

\subsection{Conversation Style Transfer using Few-Shot Learning}
\cite{conversation} introduces a few-shot learning approach for conversation style transfer, converting input conversations to match a target style using a few example dialogues. The method adopts a two-step process: first, it reduces source conversations to a style-free form via in-context learning with large language models (LLMs), then rewrites the style-free dialogue to align with the target style. This approach mitigates challenges in defining style attributes and addressing parallel data scarcity.

Human evaluations show that incorporating multi-turn context enhances style matching and improves appropriateness/semantic correctness relative to utterance- or sentence-level style transfer. Additionally, the technique proves beneficial for downstream tasks like multi-domain intent classification: transferring training data styles to match test data improves F1 scores.

A major limitation is the reliance on manually constructed style-to-style-free parallel conversations, which may be impractical for large-scale style domains. Furthermore, while increased contextual information improves appropriateness, it risks diminishing style strength and generating semantically dissimilar responses. This highlights current LLM limitations in conditioning on extensive contexts during style transfer. The study also notes suboptimal context length settings in their framework.

\section{Methods}

\begin{figure}[t]
\centering
\includegraphics[width=0.9\columnwidth]{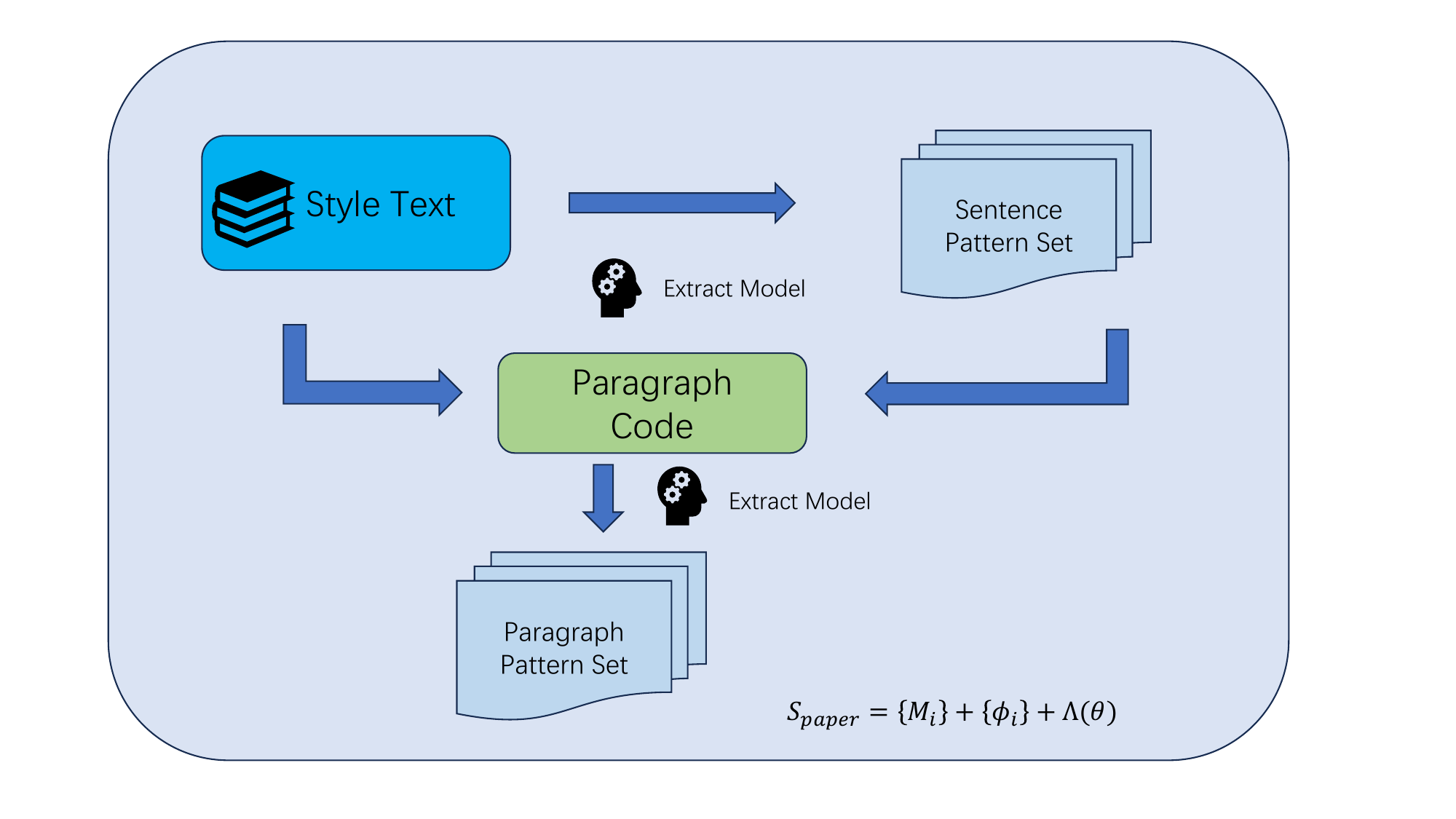} 
\caption{ZeroStylus Pipeline of Phase 1: to preprocess style text and build sentence and paragraph pattern set.}
\label{fig1}
\end{figure}

Motivated by the limitations of existing text style transfer methods, we propose \textbf{ZeroStylus}, a framework for long-text style transfer based on large language models (LLMs) zero-shot learning. This framework operates through automated semantic pattern matching without need for LLM training, while maintaining extensibility for both personal writing assistance and formal paper stylization. The algorithm accepts three inputs:
\begin{itemize}
    \item Source academic text \( T_s \)
    \item Reference papers \( \{ R_1, R_2, \ldots, R_n \} \) representing the target style
    \item Style intensity parameter \( \alpha \in [0,1] \)
\end{itemize}
producing output text \( T_o \) that preserves the source content while aligning with the rhetorical patterns of the reference papers. The primary technical challenge lies in achieving consistent style transformation across long documents, as sentence-level modifications often fail to maintain coherent stylistic patterns at the discourse level. Particularly, academic writing stylization differs massively from generic style transfer through its modular organization and structural predictability, so it's naturally a good test scene. We collect academic articles from different authors as style transfer sources from public dataset\cite{paper1,paper2}. Academic style is formally characterized through three components:

\begin{equation}
    S_{\text{paper}} = \underbrace{\{ M_1, M_2, \ldots, M_k \}}_{\text{Section Modules}} + \underbrace{\{ \Phi_1, \Phi_2, \ldots, \Phi_m \}}_{\text{Rhetorical Structures}} + \underbrace{\Lambda(\theta)}_{\substack{\text{Disciplinary} \\\text{Conventions}}}
\end{equation}

where each module \( M_i \) contains specific logical structures (e.g., literature review templates, methodology descriptions).

We employ a unified model \( \pi \) to accomplish text style transfer through two systematically coordinated phases. The architecture maintains two discrete template repositories: \( \Gamma_s \) for sentence-level patterns and \( \Gamma_p \) for paragraph-level structural features, both dynamically updated during processing.

\begin{figure}[t]
\centering
\includegraphics[width=0.9\columnwidth]{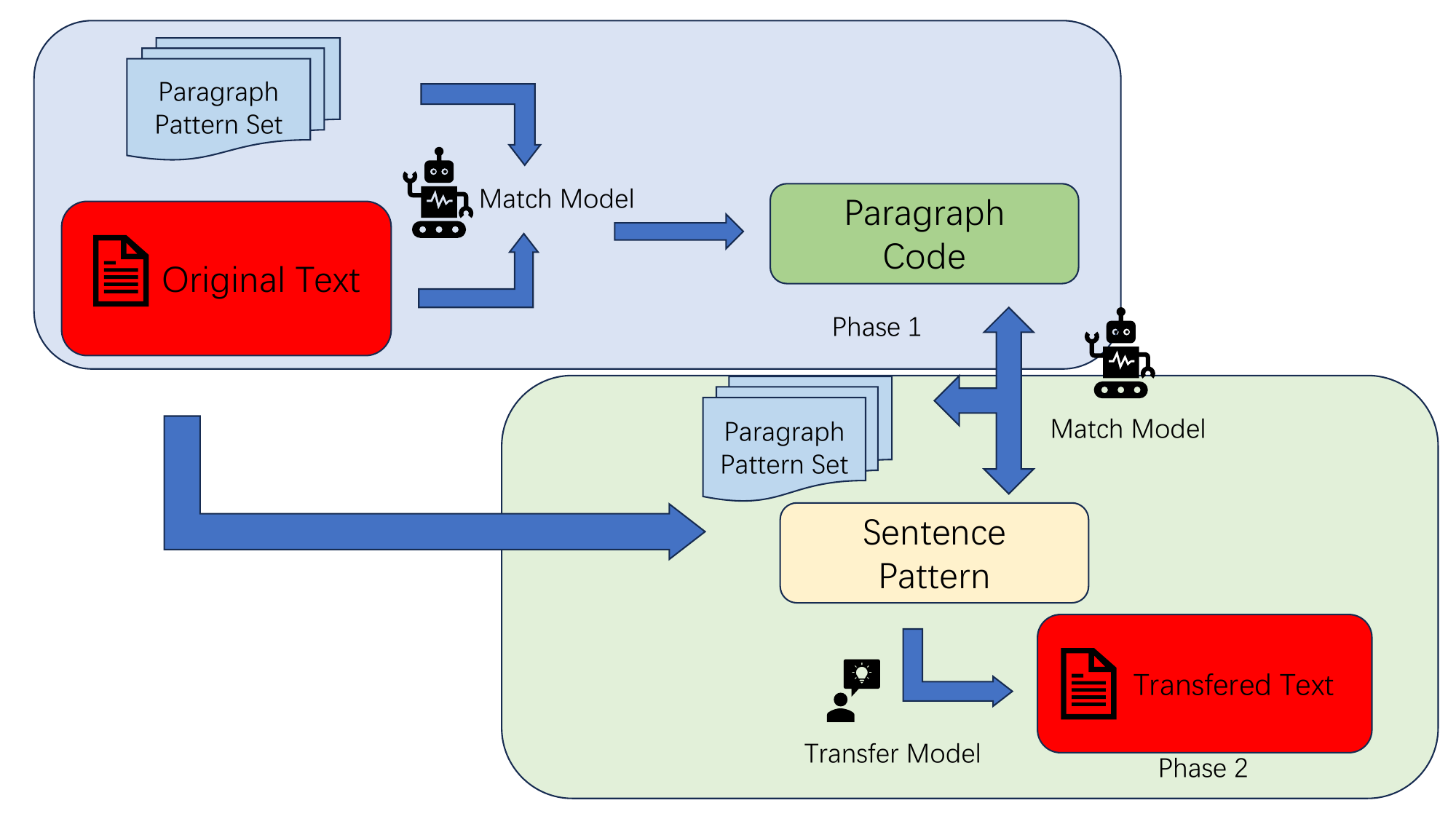} 
\caption{ZeroStylus Pipeline of Phase 2: to do transfer based on extracted pattern set in Phase 1 and input text with LLMs.}
\label{fig2}
\end{figure}

\subsubsection{Phase 1: Hierarchical Template Acquisition}

\textbf{Input:} A collection of representative text documents \( D = \{ d_1, \ldots, d_N \} \) exemplifying the target writing style.

\textbf{1.1 Sentence Pattern Extraction:} 
As shown in \ref{fig1}, extractor model processes each sentence \( s_j \) within the style corpus through its encoder component \( \pi_{enc} \), generating dense vector representations \( e_j = \pi_{enc}(s_j) \). These sentence embeddings capture latent syntactic and lexical patterns. Using density-based clustering (DBSCAN), the system identifies recurrent sentence structures by grouping embeddings with similar spatial distributions. Each cluster centroid forms a prototypical sentence template \( \tau_s \), which abstracts surface variations while preserving core stylistic elements. The resulting templates constitute the sentence repository \( \Gamma_s \), ensuring coverage of diverse expression patterns without redundant duplication.

\textbf{1.2 Paragraph Structure Modeling:} 
From below part of \ref{fig1}, for paragraph-level style analysis, the model aggregates sentence embeddings within each paragraph through hierarchical encoding: \( e_p = \pi_{enc}([ e_1, \ldots, e_m ]) \). This composite embedding captures inter-sentence relationships and discourse patterns characteristic of the target style. The paragraph template repository \( \Gamma_p \) evolves dynamically through incremental updates – a new template is added only when its embedding differs sufficiently from existing entries (\( \min_{\tau_p \in \Gamma_p} || e_p - \tau_p || > \epsilon \)). This threshold-controlled expansion prevents template proliferation while accommodating genuine structural variations and supporting continuous updates.

\subsubsection{Phase 2: Template-Guided Generation}

\textbf{Input:} Source paragraph \( p^{src} = \{ s_1, \ldots, s_n \} \) requiring style adaptation.

During this phase, we generate transferred text based on the sentence and paragraph patterns set extracted in the first phase, with following three steps as in \ref{fig2}.

\textbf{2.1 Multi-Granular Template Matching:}
The system establishes style correspondences at both linguistic and structural levels. For each source sentence \( s_i \), the encoder computes its style signature \( e_i^{src} = \pi_{enc}(s_i) \), then retrieves the closest-matching sentence template:
\begin{equation}
\tau_s^i = \underset{\tau \in \Gamma_s}{\arg\max}\ \text{sim}(e_i^{src}, \tau)
\end{equation}
Concurrently, the entire paragraph embedding \( e_p^{src} = \pi_{enc}(p^{src}) \) guides selection of the optimal structural template:
\begin{equation}
\tau_p^* = \underset{\tau_p \in \Gamma_p}{\arg\min}\ || e_p^{src} - \tau_p ||
\end{equation}
This dual matching ensures local stylistic consistency and global coherence, as displayed in blue frame in \ref{fig2}.

\textbf{2.2 Context-Aware Sentence Transformation:}
Each source sentence undergoes style infusion through the generator component:
\begin{equation}
s'_j = \pi_{gen}(s_j, \tau_s^j, \tau_p^*)
\end{equation}
In lower green frame in \ref{fig2}, the generation process simultaneously considers: (1) lexical patterns from the matched sentence template \( \tau_s^j \), (2) structural constraints from the paragraph template \( \tau_p^* \), and (3) original semantic content. This tripartite conditioning enables context-sensitive style transfer that preserves content integrity while adapting expression forms.

\textbf{2.3 Paragraph-Level Coherence Enhancement:}
The initially transformed sentences \( \{ s'_1, \ldots, s'_n \} \) are subsequently refined through structural optimization:
\begin{equation}
p^{out} = \pi_{refine}([ s'_1, \ldots, s'_n ], \tau_p^*)
\end{equation}
The refinement module adjusts inter-sentence transitions, discourse markers, and referential consistency to align with the structural template \( \tau_p^* \). This final processing step ensures the generated paragraph exhibits native-style flow and logical progression, transcending mere sentence-level style adaptation.

\section{Experiments}
\subsection{Style Text and Transfer Pipeline}
For long-text style transfer we execute following steps under specified order. Randomly sample a first author and a subset of their articles with $N_{exp}$ranging from 1 to 5, to serve as reference style text, ensuring the total length $S$ aligns with that of original text to be transferred at a rate of about $\sigma=3.0$. Next sample long-text paragraphs to be stylized, matching them with the reference articles via keyword and paper abstract based field alignment as \cite{pasa} proposed, and filtering out qualified segments unrelated to the reference authors or articles. The hierarchical framework introduced in the Methods section then performs the style transfer. Throughout this stage, we employ both GPT4-o \cite{gpt4} and DeepSeek-R1\cite{ds-r1} in parallel as the encoder, extractor, and transferer for style extraction and transformation. In following evaluation pipeline, the stylized outputs from both models are assessed independently, and their results are averaged. All subsequent method evaluations reflect this mean performance.

\subsection{Benchmarking Style Transfer Quality from Different Methods}

\textbf{Setup} Given the limited availability of objective metrics for paragraph-level style transfer, we adopt a hybrid evaluation framework inspired by preference learning and benchmark scoring protocols. Our assessment pipeline combines weighted model-based scores with human evaluations, where annotators rate stylized outputs conditioned on source paragraphs and reference style exemplars. We evaluate $N=500$ test samples randomly selected from academic paper dataset ArxivPapers and Arxiv 10, specifically introduced in \cite{paper1} and \cite{paper2}, with all scores undergoing min-max normalization before weighted fusion. Human evaluations are conducted by multiple judges, ranking from 0 (poor performance) to 10 (best performance). 

Our evaluation employs a tri-axial metric $\mathbf{v} = [x, y, z]$, where:
\begin{itemize}
    \item $x$ quantifies style consistency via paragraph-level embedding similarity between output and reference texts (computed via $\pi_{enc}$), reflecting structural alignment to $\Gamma_p$.
    \item $y$ measures content preservation by combining BLEURT scores with keyword retention recall, addressing leakage issues observed in prior template-based methods.
    \item $z$ assesses expression quality through human preference checks integrated with LLM benchmark standards (e.g., \cite{teval}) for text naturalness.
\end{itemize}

The average score is calculated as $A = \frac{X+Y+Z}{3}$.
We benchmark against four paradigm categories:
\begin{enumerate}
    \item \textbf{Original}: Unmodified input paragraphs as a control baseline.
    \item \textbf{DirectPrompt}: Simulates common zero-shot LLM usage (as in \cite{author-style}), revealing baseline performance without structural modeling.
    \item \textbf{ConvTransfer}: Implements the approach from \cite{conversation}, representing state-of-the-art sentence-level transfer.
    \item \textbf{TemplateOnly}: Our ablated variant without paragraph templates ($\Gamma_p=\emptyset$), isolating the impact of hierarchical template matching proposed in the methods section.
    \item \textbf{StructuredRewritten}: Our full approach introduced in the methods section, consisting of two phases.
\end{enumerate}

This metric design directly addresses the core challenges outlined in the introduction, balancing style strength and content integrity while ensuring linguistic naturalness.

\textbf{Result}
\begin{table*}
\centering
\caption{Evaluation result of differents style transfer methods. The best performance in model groups with size relative size is \textbf{bolded} except the reference ones.}
\label{table1}
\begin{tblr}{
  row{even} = {r},
  row{3} = {r},
  row{5} = {r},
  row{7} = {r},
  column{1} = {c},
  cell{1}{2} = {c},
  vline{2} = {-}{},
  hline{1-2,4,8} = {-}{},
}
                    & Style Consistency (X) & Content Preservation (Y) & Expression Quality (Z) & Average       \\
Original            & 4.04                  & 7.70                     & 5.83                   & /             \\
Style               & 8.10                  & 3.85                     & 6.20                   & /             \\
DirectPrompt        & 6.42                  & \textbf{7.34}            & \textbf{6.34}          & 6.70          \\
ConvTransfer        & 7.45                  & 6.22                     & 6.08                   & 6.58          \\
TemplateOnly        & \textbf{7.62}         & 5.91                     & 6.32                   & 6.62          \\
StructuredRewritten & 7.39                  & 7.04                     & 6.26                   & \textbf{6.90} 
\end{tblr}
\end{table*}

The experimental results demonstrate the following findings. As \textbf{semantic preservation} is measured against the original unstylized text and \textbf{stylistic similarity} against reference stylized texts, the original text naturally achieves the highest semantic preservation but lowest stylistic similarity. Conversely, the reference text exhibits the highest stylistic similarity at the expense of semantic preservation. Among comparative methods:

\textbf{DirectPrompt}, which employs complete unstylized text and reference style prompts for holistic stylization, achieves superior semantic preservation but the lowest stylization degree. This results from LLMs’ tendency to partially stylize only initial paragraphs while minimally modifying subsequent content when processing lengthy texts, yielding outputs indistinguishable from the original. 

In contrast, \textbf{TemplateOnly}, which performs sentence-level stylization by jointly inputting sentences with reference style texts, achieves higher stylization scores but suffers significant semantic degradation. This stems from its strict imitation of reference sentence patterns without modeling inter-sentential logical relationships (e.g., progression or parallelism), thereby disrupting structural coherence despite improved stylization coverage.

\textbf{ConvTransfer}, which processes multi-turn dialogues via destylization and restylization of individual utterances, exhibits similar limitations to \textbf{TemplateOnly}. While achieving comparable stylization through per-sentence processing, it loses contextual structural information during destylization, though this is partially mitigated by multi-sentence batch processing. 

Our proposed \textbf{StructuredRewritten} combines hierarchical paragraph-level template matching with sentence-level rewriting, preserving \textbf{TemplateOnly}'s stylization strength while maintaining \textbf{DirectPrompt}'s paragraph-level semantic coherence. Notably, all methods achieve similar human preference scores exceeding the original text, probably due to shared LLM alignment strategies that enhance social preference conformity.

\subsection{Adversarial Evaluation}
To rigorously assess macroscopic style persistence, we implement a pairwise comparative framework that directly evaluates structural coherence capabilities across methods. The evaluation pipeline comprises three components:

\textbf{Input Tuple:} 
\begin{equation}
\mathcal{I} = (p^{src}, p^{ref}, p^{out}_A, p^{out}_B) \in \mathbf{P}^4
\end{equation}
where $p^{src}$ denotes the source paragraph, $p^{ref}$ the style reference, and $\{p^{out}_A, p^{out}_B\}$ outputs from competing methods.

\textbf{Evaluation Process:}

1. \textit{Model Prompting}: For each evaluator model $M \in \{\text{GPT-4o}, \text{DeepSeek-R1}, 
\text{Llama-4}\}$ \cite{gpt4,llama,ds-r1}, generate preference scores using standardized prompts:

\begin{align}
s_M^{(A,B)} &= f_M(\langle p^{src}, p^{ref}, p^{out}_A\rangle) \\
&= f_M(\langle p^{src}, p^{ref}, p^{out}_B\rangle)
\end{align}

2. \textit{Position Bias Mitigation}: Compute positional-robust preference scores:
\begin{equation}
\text{Pref}_M(A) = \frac{1}{2}\left[\sigma(s_M^{(A,B)}) + (1-\sigma(s_M^{(B,A)}))\right]
\end{equation}
where $\sigma$ denotes the sigmoid normalization function.

3. \textit{Aggregate Winning Rate}: For method $\pi$ against baseline $\beta$ across $N$ samples:
\begin{equation}
\text{WinRate}(\pi) = \frac{1}{N}\sum_{i=1}^N \mathbf{I}\left[\text{Pref}_M(\pi_i) > 0.5 + \delta\right]
\end{equation}
with $\delta=0.1$ as the decision margin to account for model uncertainty.

We conduct adversarial evaluations between \textbf{TemplateOnly} and \textbf{SentencePattern} (which extracts only sentence patterns) to demonstrate the effectiveness of pattern set extraction. Additionally, we compare \textbf{SentencePattern} with \textbf{StructuredRewritten} (which extracts both sentence patterns and paragraph index patterns, initially matching paragraph patterns) to highlight the advantage of preserving layered style during transfer. For $N_1=100$ samples, we report win-or-lose percentages between competing methods.

\textbf{Result}
\begin{table}
\centering
\caption{Adversarial Evaluation between TemplateOnly and SentencePattern Methods: Win Rate}
\label{table2}
\begin{tblr}{
  row{even} = {r},
  row{3} = {r},
  column{1} = {c},
  cell{1}{2} = {c},
  vline{2} = {-}{},
  hline{1-2,5} = {-}{},
}
{SentencePattern \\vs TemplateOnly} & {GPT\\4o\\(\%)} & {Ds\\-R1\\(\%)} & {Llama\\-4\\(\%)} \\
{Style Consistency\\~(X)~}          & 56              & 55              & 61                \\
{Content Preservation\\~(Y)~ ~}     & 53              & 53              & 54                \\
{Expression Quality\\~(Z)}          & 50              & 52              & 51                
\end{tblr}
\end{table}

The first ablation study (\textbf{TemplateOnly} vs. \textbf{SentencePattern}) reveals that \textbf{SentencePattern}'s pre-extracted deduplicated sentence templates significantly enhance style transfer accuracy (57.3\% win rate in average) while marginally improving semantic preservation (53.3\% win rate in average). This improvement stems from reduced template mismatch errors and minimized leakage of non-stylistic details from reference texts. Furthermore, comparable human preference scores indicate limited impact on alignment quality.

\begin{table}
\centering
\caption{Adversarial Evaluation between SentencePattern and StructuredRewritten Methods: Win Rate}
\label{table3}
\begin{tblr}{
  row{even} = {r},
  row{3} = {r},
  column{1} = {c},
  cell{1}{2} = {c},
  vline{2} = {-}{},
  hline{1-2,5} = {-}{},
}
{TemplateOnly vs \\StructuredRewritten} & {GPT\\4o(\%)} & {Ds\\-R1(\%)} & {llama\\-4(\%)} \\
{Style Consistency \\(X)~}              & 52            & 54            & 55              \\
{Content Preservation \\(Y)~ ~}         & 46            & 39            & 44              \\
{Expression Quality \\(Z)}              & 48            & 53            & 50              
\end{tblr}
\end{table}

In the second ablation group (\textbf{SentencePattern} vs. \textbf{StructuredRewritten}), our two-stage framework keeps close in stylization strength (> 46\% win rate) while significantly improving semantic preservation (57\% vs 43\% win rate) through paragraph-level structural encoding. This validates that hierarchical template matching better preserves inter-sentence relationships compared to pure sentence-level processing. At the same time in the expression quality dimension two methods are tightly grasped with around 50\% in all the win-or-lose samples, onfirming that structural encoding does not degrade text alignment quality.

\section{Discussions}
Although experiments demonstrate novel framework's effectiveness compared to strict zero-shot baselines, several limitations remain, prompting directions for future work.

\begin{itemize}
\item Benchmarking Long-Text Style Transfer in further systematical manner. Current benchmarks for dialogue or paragraph-style transfer lack systematic quantitative evaluation capabilities for long-form articles, highlighting the need for dedicated metrics.
\item Semantic Splitting for Rewriting. Replacing basic period-based splitting with sentence-level semantic segmentation could better isolate cross-sentence independent semantics, improving unit-level rewriting and semantic capture.  
\item Style-Specific Evaluation. Author-style assessments may vary significantly across domains and applications, necessitating task-specific template extraction and tailored evaluation of matching effects.  
\item Hierarchical Semantic Parsing. The two-layer framework could be further extended, including incorporating paragraph-level features e.g., types, roles, inter-paragraph relationships, to enable structured article encoding and semantic re-layout, while further extensions might include systematic structural design across documents or code files.  
\end{itemize}

\section{Conclusion}
We introduce ZeroStylus, a zero-shot framework for long-text style transfer that addresses key limitations in current LLM-based approaches through hierarchical template matching. By decoupling sentence-level pattern extraction from paragraph-level structural modeling, our method achieves better content preservation and style consistency score while maintaining relative overall quality metrics compared to baselines, outperforming conventional sentence-level transfer approaches. The two-phase architecture demonstrates that explicit encoding of rhetorical structures combined with dynamic template repositories effectively mitigates style drift in extended text generation. Experimental validation across multiple paradigms confirms the framework's ability to preserve both micro-stylistic features and macro-structural patterns, with adversarial tests showing  preference over ablated variants. Future work should explore multilingual adaptation and efficient template updating mechanisms to enhance applicability across diverse stylistic domains. Our findings suggest that hierarchical style representation with constrained-context rewriting offers a viable pathway for coherent long-text transformation in resource-constrained scenarios.

\bibliography{aaai22}

\end{document}